# Mobile Robot Yielding Cues for Human-Robot Spatial Interaction*

Nicholas J. Hetherington†, Ryan Lee†, Marlene Haase, Elizabeth A. Croft, H.F. Machiel Van der Loos

*Abstract*—Mobile robots are increasingly being deployed in public spaces such as shopping malls, airports, and urban sidewalks. Most of these robots are designed with human-aware motion planning capabilities but are not designed to communicate with pedestrians. Pedestrians encounter these robots without prior understanding of the robots' behaviour, which can cause discomfort, confusion, and delayed social acceptance. In this research, we explore the common human-robot interaction at a doorway or bottleneck in a structured environment. We designed and evaluated communication cues used by a robot when yielding to a pedestrian in this scenario. We conducted an online user study with 102 participants using videos of a set of *robot-to-human yielding cues*. Results show that a Robot Retreating cue was the most socially acceptable cue. The results of this work help guide the development of mobile robots for public spaces.

## I. INTRODUCTION

Mobile robots are well accepted in controlled environments but are increasingly being deployed in public spaces such as shopping malls, airports, and urban sidewalks. The International Federation of Robotics predicts a 323% increase in mobile robot use for logistics from 2018 to 2021 [1]. McKinsey & Co. predicts mobile robots will complete 80% of last-mile deliveries in the future [2]. These robots benefit their owners and end users but can be disruptive when poorly designed. We focus on the design of mobile robots for *human-robot spatial interaction* (HRSI) in public spaces with pedestrians.

Human-human interaction while walking in public spaces is natural but relies on an understanding of how others move [3]. Furthermore, natural crowd movement relies on pedestrians' subtle body language cues [4]. Mobile robots are new, unfamiliar agents in public spaces. Delivery and logistics robots are designed to move safely and efficiently but do not communicate with pedestrians apart from simple visual/audio cues. As a result, more pedestrians are encountering spatial interactions with these robots without prior understanding of their behaviour.

Most HRSI research focuses on human-aware navigation and motion planning, such as designing robot motion that adheres to social conventions in public spaces. For example, Chen *et al.* proposed a method for "inducing socially aware behaviours in a reinforcement learning framework" [5]. Trautman *et al.* proposed Interacting Gaussian Processes as "the first algorithm that explicitly models human cooperative collision avoidance for navigation in dense human crowds" [6]. In their survey, Kruse *et al.* presented three goals of human-aware navigation systems: human comfort, robot naturalness, and robot sociability [7]. However, most approaches to human-aware navigation do not include robot legibility. Misunderstanding robots' behaviour can lead to discomfort [8] and jeopardize safety [9]. In addition to human awareness, legible robot behaviour is a key component of HRSI [10]. Lasota *et al.* wrote that "human agents' ability to predict the actions and movements of a robot is as essential as the ability of a robot to predict the behaviour of humans" [11]. There are a variety of definitions for the legibility of robot behaviour in the literature. We use the broad definition from Lichtenthäler *et al.* that "robot behaviour is legible if a human can infer the next actions, goals and intentions of the robot with high accuracy and confidence" [10]. Legible robot behaviour can help achieve Kruse *et al.*'s three goals of comfort, naturalness, and sociability.

In this paper, we consider a common HRSI in public spaces: namely, at a *doorway* or similar narrowed passage. This situation inherently contains conflict because only one agent at a time can proceed through the narrow space, and the pedestrian may be uncertain whether they or the robot should go first. We developed and evaluated *robot-to-human yielding cues*, which communicate a mobile robot's intent to yield to a pedestrian. Our specific contribution identifies which robot yielding cues are communicative in this context, providing insight for designing these human-robot spatial interactions.

## II. RELATED WORK

Multiple works investigate human-robot head-on interactions in hallways or corridors in which the robot passes a human in a narrow space. In a proxemics study, Lauckner *et al.* established minimal frontal and lateral distances for a mobile robot approaching a person in a corridor. Participants teleoperated the mobile robot and drove it towards themselves until they felt uncomfortable [12]. Dondrup *et al.* showed that the robot's deceleration within the pedestrians' personal space resulted in less disruption to their movement [13]. The corridor context in these studies is slightly different than the *doorway* context in our research, but the results can inform the development of robot yielding cues with appropriate human-robot proxemics and robot velocities.

*Supported by the Natural Sciences and Engineering Research Council of Canada, Government of British Columbia, and Australian Research Council Discovery Project DP200102858. N.J. Hetherington, R. Lee, and H.F.M. Van der Loos are with the Department of Mechanical Engineering, University of British Columbia, Vancouver, BC, Canada (nickjh@alum.ubc.ca, ryan.lee@alumni.ubc.ca, vdl@mech.ubc.ca). M. Haase is with Human Factors Engineering, Technical University of Munich (marlene.schlorf@tum.de). E.A. Croft is with the Departments of Mechanical and Aerospace Engineering and Electrical and Computer Systems Engineering, Monash University, Melbourne, VIC, Australia (elizabeth.croft@monash.edu).

† These authors contributed equally.

A similar and common context in the autonomous road vehicle (AV) literature addresses an AV yielding to a pedestrian at a crosswalk. Ackermann *et al.* showed that smooth and early AV deceleration decreased the time pedestrians took to decide whether the vehicle was yielding to them [14]. Interestingly, the AV literature often focuses on communicating that the AV will *not* yield to the pedestrian (e.g. Gupta *et al.* [15]). Accordingly, Thomas *et al.* developed a mobile robot with "assertive" behaviour. The assertive behaviour used high acceleration to show pedestrians that the robot would *not* yield to them at a doorway in a head-on interaction [16]. These works demonstrate that acceleration and deceleration can be used as a robot-to-human yielding cue.

Retreating or diverting is a common yielding cue for mobile robots in the literature. Kaiser *et al.* compared communication cues for a robot to yield to a pedestrian at a narrow passage in two scenarios: one where the robot and the human approached the narrow passage side by side, and another where they approached from opposite directions. In each scenario, they compared pre-emptive stopping and moving aside to the robot proceeding through the narrow passage without yielding. In both scenarios, they found both cues to increase the robot's behaviour legibility but moving aside was more legible of the two [17]. In addition to their assertive motion planner, Thomas *et al.* developed a yielding cue in which the robot moved aside before the doorway [16]. In a spatial interaction between a human and a mobile robot looking at the same object, Akita *et al.* developed an approach to divert the robot to an optimal position for both parties, resulting in fewer collisions than a human-agnostic approach [18]. Retreating has also been used as a robot yielding cue in literature on human-robot collaboration with articulated robot arms. Moon *et al.* first showed that an articulated robot arm can use human-inspired hesitation to communicate yielding in a collaborative picking task [19]. In a collaborative pick and place task, Reinhardt *et al.* showed a stop-then-retreat cue increased trust compared to a dominant, non-yield cue [20].

The literature suggests that robot deceleration, retreating, or diversion are effective robot-to-human yielding cues. Furthermore, no studies reviewed here have investigated the particular HRSI shown in Fig. 1, a head-on interaction in which both agents wish to enter a doorway to their side.

### III. DESIGN AND FUNCTION OF YIELDING CUES

For the HRSI in Fig. 1, we designed and prototyped five robot-to-human yielding cues: (1) Stop, (2) Decelerate, (3) Retreat, (4) Tilt, and (5) Nudge. Fig. 2 shows the cues. In each cue, the robot moves towards the doorway with a linear speed of *v* before performing a movement to communicate that the pedestrian should enter first. The paragraphs below describe the yielding cues in detail. We conducted an in-person pilot study with five participants to set the values for these parameters (Table 1).

Table 1. Cue parameter values.

| Cue | Parameter | Units | Value |
|---|---|---|---|
| All | Linear speed, $v$ | m/s | 0.7 |
| All | Angular speed, $\omega$ | °/s | 60 |
| Decelerate | Distance, $X_d$ | m | 1 |
| Retreat | Distance, $X_r$ | m | 0.1 |
| Tilt | Angle, $\theta_t$ | degrees | 40 |
| Nudge | Angle, $\theta_n$ | degrees | 55 |
| Nudge | Number, $N_n$ | - | 2 |

In the Stop cue, the robot stops abruptly at the doorway edge and is used as a control condition in this study. Many reviewed studies use a similar stopping cue as a control (e.g. Kaiser *et al.* [17] and Moon *et al.* [19]).

In the Decelerate cue, the robot begins to decelerate at a distance of $X_d$ before stopping at the doorway edge. In the Retreat cue, the robot stops abruptly at the doorway edge, then retreats a distance of $X_r$ and stops. The literature reviewed above shows that retreating cues (e.g. Reinhardt *et al.* [20]) and decelerating cues (e.g. Ackermann *et al.* [14]) are effective robot yielding cues, but they have not been directly compared.

In the Tilt cue, the robot stops abruptly at the doorway edge, then turns away from the doorway at an angular speed of ω until an angle of $\theta_t$. The Tilt cue is evocative of pedestrians turning away from a doorway and is similar to a diversion cue that literature had shown to be an effective yielding cue.

In the Nudge cue, the robot stops abruptly at the doorway edge, then makes $N_n$ rotations, or "nudges", towards the doorway to an angle of $\theta_n$ with an angular speed of $\omega$. The robot cue rotates back to its approach angle between each "nudge" and finishes facing the pedestrian. The Nudge cue is evocative of a pedestrian waving their hand or turning their head when yielding to another pedestrian at a doorway.

Tilt and Nudge are both anthropomorphic yielding cues. We designed Tilt and Nudge to test whether anthropomorphic cues could be communicative in our scenario (Fig. 1).

We implemented the robot yielding cues on a PowerBot differential drive mobile base (Adept Mobile Robots, Amherst, NH, USA) using the Robot Operating System. The mobile base is 83 cm long, 63 cm wide, and 49 cm tall with an added sensor tower (total 173cm). To facilitate rapid prototyping and simple evaluation, we implemented each cue with open-loop velocity commands specific to the scenario and for demonstration purposes only.

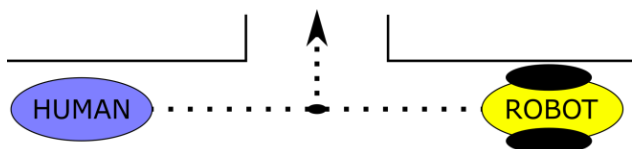

Fig. 1: Diagram of the interaction investigated in this research, showing the agents' intended paths.

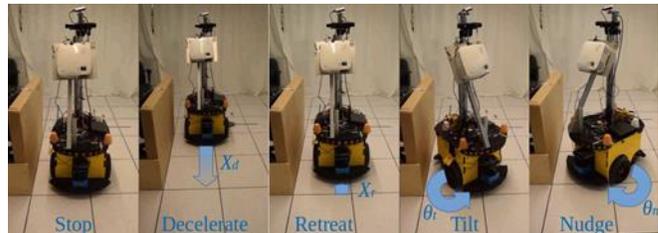

Fig. 2. Robot yielding cues shown from the human's perspective in Fig. 1. The doorway is to the robot's right. The images show the point at which the robot starts the Decelerate ($X_d$), the point to which the robot reverses in Retreat ($X_r$), and the rotational maxima for Tilt and Nudge ($\theta_t$ and $\theta_n$).

## IV. USER STUDY

We conducted an online user study with 102 participants using videos of the robot yielding cues and a digital survey. The user study was approved by the University of British Columbia's Behavioural Research Ethics Board. A video describing the user study is attached as supplementary material. The original videos are archived in [21].

### A. Outcome Measures and Hypotheses

We designed the user study to test the following hypotheses, that the yielding cues are statistically different in terms of: (H1) trustworthiness; (H2) comprehensibility; (H3) participants' comfort during the interaction; (H4) likeability; and (H5) social compatibility. These five measures are all components of social acceptability. We also measured participants' interpretation of the robot yielding cues.

### B. Experimental Scenario

We captured a video of each cue from the human perspective shown in Fig. 1. As shown in the accompanying video, we applied a zooming effect to simulate the viewer walking towards the doorway. We used a zooming effect for more consistent videos, rather than introducing variance from human walking. The videos were silent to avoid confusion and eliminate sound as a confounding factor. We tested three other video styles using an online pilot test with 29 participants, including a static video and a moving red dot or stick figure to represent a human walking towards the doorway. The zooming animation style was preferred by the participants.

### C. Survey Design and Experimental Procedure

After giving consent, participants read instructions explaining the situational context of the videos, accompanied by a diagram of the scene. We did not state that all the cues were designed to communicate yielding. Thus, participants had to interpret each cue for themselves. We used only yielding behaviours; a non-yielding condition would have introduced the robot's intention as a confounding factor. Participants first viewed a familiarization video of the robot entering the doorway without making a cue. The next five videos were presented in a randomized order to all participants to show each robot yielding cue: Stop, Decelerate, Retreat, Tilt, and Nudge.

After each video, the survey first asked participants an attention check question: 1. "Did the robot rotate during the video, either to its left or right? A) Yes, or B) No." We excluded participants who answered Question 1 incorrectly. The survey then asked participants about their interpretation of the yielding cue: 2. "According to the robot's movement cue, should the robot or you (the viewer) enter the doorway first? A) Robot, or B) Me."

Participants then responded to the following statements on a 7-point Likert scale from "Very Strongly Disagree" to "Very Strongly Agree":

- 3.1. "I was confident in deciding who should enter the doorway first."
- 3.2. "The robot's movement cue was misleading for me."
- 3.3. "I quickly understood the robot's movement cue."
- 3.4. "The robot's movement cue was sufficient for me to decide who should go through the door first."
- 3.5. "It is difficult to understand what the robot does."
- 3.6. "I trust the robot."
- 3.7. "I can rely on the robot."
- 3.8. "The robot is deceptive."
- 3.9. "I am wary of the robot."
- 3.10. "The robot's movement cue would be socially compatible in a pedestrian's environment."
- 3.11. "The robot's movement cue made me feel comfortable."
- 3.12. "I liked the robot."

We used two Likert scales constructed from the statements in Question 3. We used a Cue Comprehension Scale constructed from Statements 3.2-3.5, and a Cue Trust Scale constructed from Statements 3.6-3.9. Each scale has two statements with positive valence and two with negative valence. We calculated the score for each scale as the mean response to the statements on a 7-point numeric scale. Joshi *et al.* support taking the mean of a set of Likert-type items to create interval data [22]. The comprehension items are partially adapted from Körber's Trust in Automation Questionnaire [23] and Madsen's Human-Computer Trust Scale [24]. Trust items 3.6 and 3.7 are adapted from Körber's Trust in Automation Questionnaire and trust items 3.8 and 3.9 are adapted from Jian *et al.*'s System Trust Scale [25]. We informally pilot tested both scales with 7 participants to verify that their understanding of the statements matched ours.

### D. Participants

We recruited participants using Amazon Mechanical Turk for a reward of US $2.50. Out of a total of 128 responses to the online survey, 25 were excluded due to incorrect responses to the attention check and one was excluded due to a self-reported misinterpretation, leaving 102 responses to analyze. The participant excluded for misinterpreting was paid, but the ones excluded for failing the attention check question were not.

Of the 102 participants, 31 identified as female, 70 identified as male, and 1 preferred not to say. They also placed themselves in age brackets: 20% were ages 18-25, 21% were 26-30, 34% were 31-40, 15% were 41-50, 8% were 51-60, and 2% were over 60. Participants also described their prior experience with robots, for which none were excluded.

### E. Data Analysis

We analyzed the internal reliability of the Cue Likert scales for each yielding cue condition using Cronbach's $\alpha$ and found them satisfactory by the general convention of 0.7.

We performed repeated measures ANOVA tests on the comprehension and trust data. Because we had a large sample size (N = 102) we could ignore the assumption of normality for the repeated measures ANOVA test [26]. Furthermore, there were no extreme outliers in the data. Neither the comprehension data nor the trust data met the assumption of sphericity, so we used the Greenhouse-Geisser correction. We also calculated the Pearson correlation coefficient between the trust and the comprehension of each robot yielding cue. These analyses tested hypotheses H1 and H2.

We performed exploratory Friedman's ANOVA tests on each of the Likert items 3.10 (social compatibility), 3.11 (comfort), and 3.12 (likeability). We used the non-parametric Friedman's ANOVA because these data are on ordinal scales, whereas the comprehension and trust data are on interval scales. These analyses tested hypotheses H3-H5. We performed Bonferroni-corrected post-hoc tests on the significant effects detected by both the parametric and non-parametric ANOVA tests (H1-H5).

We analyzed Question 2 in conjunction with Statement 3.1 to explore how correctly and confidently participants interpreted the cues as yielding or not.

## V. RESULTS

### A. Trust and Comprehension

The repeated measures ANOVA on cue trust detected a significant main effect of cue condition with a small effect size ($F(3, 292) = 3.8$, $p = .012$, $\eta_p^2 = .04$). Pairwise comparisons showed the Retreat cue was significantly more trustworthy than both the Stop cue ($M = 0.35$, $p = .002$) and the Decelerate cue ($M = 0.35$, $p = .003$). There were no significant differences in trust between other combinations of cues.

The repeated measures ANOVA on cue comprehension (hypothesis 2) detected a significant main effect of cue condition with a medium effect size ($F(3, 326) = 8.8$, $p < .001$, $\eta_p^2 = .08$). Pairwise comparisons showed the Retreat cue was significantly more comprehensible than the Stop cue ($M = 0.8$, $p < .001$), the Decelerate cue ($M = 0.7$, $p < .001$), and the Tilt cue ($M = 0.5$, $p = .029$). The Nudge cue was significantly more comprehensible than the Stop cue ($M = 0.6$, $p = .010$). There were no significant differences in comprehension between other combinations of cues.

Fig. 3 illustrates the cue trust and comprehension results. Hypotheses H1 and H2 are partially supported. There were statistically significant differences between some cues, but not all. Overall, Retreat was the most trustworthy and comprehensible.

### B. Comfort, Likeability, and Social Compatibility

The three Friedman's ANOVA tests detected a significant main effect of yielding cue on comfort ($\chi^2(4) = 27.2$, $p < .001$), likeability ($\chi^2(4) = 27.2$, $p < .001$), and social compatibility ($\chi^2(4) = 23.0$, $p < .001$). Post-hoc tests showed the Retreat cue was significantly more comfortable than the Stop cue ($p < .001$), the Decelerate cue ($p = .015$), and the Tilt cue ($p = .004$). The Retreat cue was significantly more likable than the Stop cue ($p = .005$) and the Tilt cue ($p = .049$). The Retreat cue was significantly more socially compatible than the Stop cue ($p = .019$), the Decelerate cue ($p = .035$), and the Tilt cue ($p = .003$). There were no other significant differences in comfort, likeability, or social compatibility between yielding cues.

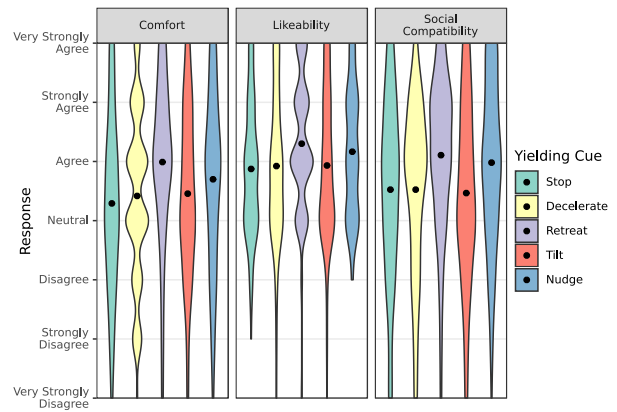

Fig. 4. Violin plots showing the responses to Statements 3.10 – 3.12. Black points show the means of the responses on a 7-point numeric scale to help visually distinguish the violins.

Fig. 4 illustrates these results. Hypotheses H3-H5 are partially supported. There were statistically significant differences between some cues, but not all. Overall, Retreat was the most comfortable, likeable, and socially compatible.

### C. Interpretation of Robot Yielding Cues

Question 2 asked participants whether they interpreted the robot's cue as yielding or not. Statement 3.1 then assessed their confidence in their interpretation. Participants correctly interpreted the cues as yielding at least 75% of the time (Stop, Nudge). Retreat was correctly interpreted the most often (85%). Stop had the lowest confidence while Retreat and Nudge had the highest confidence scores. With all cues, participants were slightly less confident deciding the robot was not yielding to them. Fig. 5 illustrates the results.

The results show that Retreat is the most socially acceptable robot yielding cue. The Retreat cue was rated higher than the Nudge cue in all five social acceptability measures, but not with statistical significance. Fig. 5 shows that Retreat was the most correctly interpreted yielding cue (85%). The Retreat cue was rated significantly higher than the

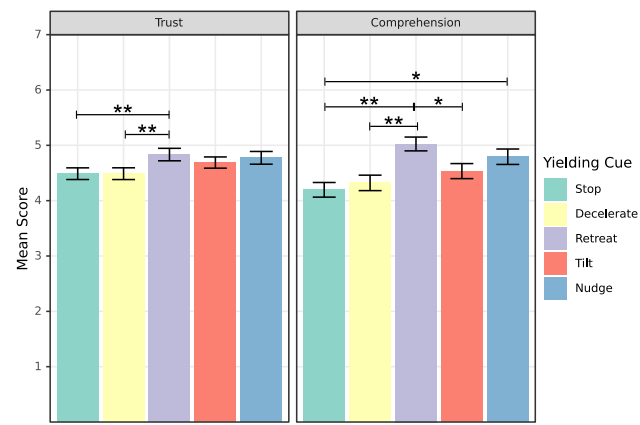

Fig. 3. Trust and comprehension scales for each yielding cue. 95% confidence intervals are shown around each mean. Brackets indicate the statistical significance of pairwise differences: **, $p < .01$; *, $p < .05$.

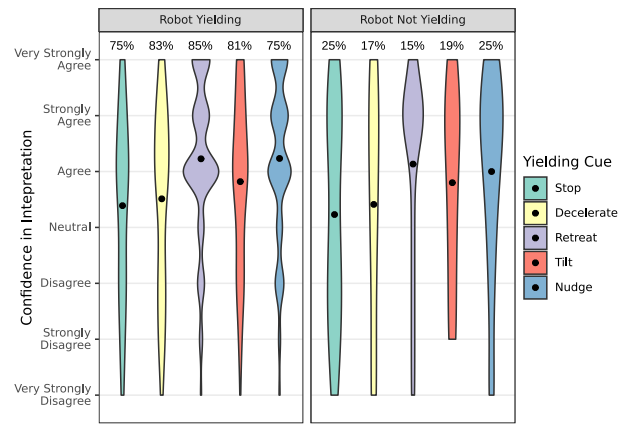

Fig. 5. Left and right graphs show the interpretation of each yielding cue with annotated percentages. Violins show participants' confidence in their interpretations. Black points show the means of the responses on a 7-point numeric scale to help visually distinguish the violins.

Stop, Decelerate, and Tilt cue in all social acceptability measures except likeability for Decelerate and trust for Tilt. Further details on methods and results can be found in [21].

## VI. DISCUSSION

Recall that hypotheses H1-H5 consider components of social acceptability. We expected the Nudge cue to be the most socially acceptable because it is the richest and the most human-like. The success of the Retreat cue does not meet this expectation. However, Nudge was rated slightly higher than the Stop, Decelerate, and Tilt cues in all five measures. This indicates that an anthropomorphic Nudge cue could be more socially acceptable than deceleration or stopping cues. Interpretations of Nudge were the most polarized and were incorrect 25% of the time but had the second-highest confidence. Adding an anthropomorphic head could help participants understand the Nudge cue.

Our results show that the previously unexplored retreating cues are more socially acceptable than deceleration and rotating cues. The results also show a strong correlation between trust and comprehension of robot yielding cues. This underlines a major motivation for this research: robots should be designed for comprehension in order to engender trust. Kauppinen *et al.* validated the relationship between trust and comprehension with the Human Computer Trust Rating Scale, which was tested with air traffic control systems [27]. We have demonstrated this relationship in robot yielding cues.

The literature shows retreating to be an effective robot yielding cue with robot manipulators (e.g. Moon *et al.* [19] and Reinhardt *et al.* [20]). Our results show that retreating can be used as an effective mobile robot yielding cue in the context explored in this paper. Recently, Reinhardt *et al.* also explored yielding cues in a similar context to ours and found support for a retreating yielding cue. They compared their retreating cue to cues similar to our Stop and Tilt, as well as to a combination of retreating and tilting. Reinhardt *et al.*'s concurrent result underlines our result and shows further support for retreating cues in HRSI [28].

In the Retreat cue design, set from pilot testing, the robot retreated a distance of 10 cm from the edge of the doorway. This is much smaller than the 45 cm distance Lauckner *et al.* found in a robot proxemics study as the minimum for human comfort [12]. Lauckner *et al.* used a head-on human-robot interaction in a corridor to identify this distance, but without a doorway to the side. The space and context afforded by the doorway in our study could have made participants feel more comfortable.

Unexpectedly, the Decelerate cue scored similarly to the Stop cue in all social acceptability measures. Fig. 5 shows that the Decelerate cue was interpreted correctly more often and with higher confidence than the Stop cue, but we did not statistically test these data. We pilot tested different distances from the doorway at which to start the deceleration ($X_d$) but not different deceleration values. Dondrup *et al.* showed that lower mobile robot velocities within a pedestrian's personal space resulted in less disruption to the pedestrian's movement [13]. It is possible a higher deceleration rate and a longer period of lower velocity would have better distinguished our Stop and Decelerate cues.

Our experimental scenario represents a stereotypic and pertinent interaction with mobile robots in public spaces. Most public spaces include some structure that creates intersections, including narrow corridors in buildings, corners on sidewalks, furniture in cafes, and cubicles in offices. We framed the experimental scenario as an interaction at a doorway to help participants immerse themselves in the scenario, in addition to piloting different immersive video styles. Despite limiting the scope of the interaction to a doorway, we believe the results translate to many other important interactions in structured public spaces. It is important to consider both the structural context and the human agent's intentions when selecting a robot yielding cue. Both our study and that in [28] found a retreating cue to communicate yielding, but in both interactions the robot retreating did not block the human's desired path. In both [16] and [17] the robot retreating backwards would have blocked the human's desired path, so the robot moved to the side. Bolstered by our results, the literature supports retreating or diverting yielding cues, depending on the structural context.

### A. Limitations

Experimentation with videos and an online survey is a limitation of the user study. The videos were not entirely representative of an in-person interaction with the robot; measures such as comfort and social compatibility are likely to be different in in-person interactions. However, there is evidence that experimentation with videos produce meaningful results. Woods *et al.* showed video results were equivalent to in-person results in a study about which direction was the most appropriate for a robot to approach a human [29]. Furthermore, Lichtenthäler *et al.* used videos to test navigation algorithms in a human-robot path crossing scenario [10]. Although the survey had an attention check, data from Amazon Mechanical Turk respondents are less trustworthy than data collected in-person due to anonymity. Furthermore, the participants were likely from a variety of cultures that may interpret communication cues differently. All the researchers and pilot test participants lived and worked in western cultures. Lastly, the survey may have been misinterpreted as we did not directly assess participants' English language skills.

## VII. CONCLUSION

In this research we designed and evaluated five different robot-to-human yielding cues for HRSI at intersections in structured environments: Stop, Decelerate, Retreat, Tilt, and Nudge. The Retreat cue scored highest in measures of trust, likeability, comprehension, comfort, and social compatibility. Retreat was correctly interpreted the most often and with the highest confidence. The doorway scenario represents a common interaction and thus makes the results applicable to many public spaces.

### A. Future Work

These results should be validated with in-person experiments, which could provoke different reactions to the yielding cues. To evaluate their robot's communication cues, some related studies collected participants' reaction or decision-making time (e.g. Ackermann *et al.* [14]), and some analyzed subjects' walking trajectories (e.g. Watanabe *et al.* [30]). Second, the Cue Comprehension and Cue Trust Likert scales should be rigorously validated with data from further studies.

Our research focuses on a common interaction in a structured context, not open spaces. Open spaces are an important but separate spatial context for interactions. To arrive at more general design guidelines, future work should explore robot yielding cues in open spaces, potentially comparing retreating and diverting cues.

Neither in this research, nor in the literature reviewed above, have robot yielding cues been designed to communicate to multiple pedestrians. How would pedestrians perceive the cues if there were other pedestrians in the scenario? Would the robot need to specify to which pedestrian it was communicating? The Retreat cue was the most socially acceptable cue in our head-on scenario, but how would pedestrians *behind* the robot react to its retreating? To answer these questions, the yielding cues should be tested further with in-person experiments.